\documentclass[sigconf]{acmart}
\AtBeginDocument{%
  }

\usepackage{enumitem}
\usepackage[table,xcdraw]{xcolor}
\usepackage{booktabs}
\usepackage{arydshln}
\usepackage{multirow}
\usepackage{graphicx}
\usepackage{pifont}
\usepackage[normalem]{ulem}
\useunder{\uline}{\ul}{}

\newcommand{\textremarkright}[1]{\textcolor{darkgreen}{\textbf{{#1}}}}
\newcommand{\textremarkwrong}[1]{\textcolor{darkred}{\textbf{{#1}}}}
\newcommand{\textremarkquestion}[1]{\textcolor{darkblue}{\textbf{{#1}}}}
\newcommand{\textremarkrepeat}[1]{\textcolor{darkyellow}{\textbf{{#1}}}}
\definecolor{darkgreen}{rgb}{0.0, 0.45, 0.0}
\definecolor{darkred}{rgb}{0.5, 0.0, 0.0}
\definecolor{darkblue}{rgb}{0.0, 0.0, 0.5}
\definecolor{darkyellow}{rgb}{0.65, 0.65, 0}

\copyrightyear{2026}
\acmYear{2026}
\setcopyright{cc}
\setcctype{by}
\acmConference[WWW '26]{Proceedings of the ACM Web Conference 2026}{April 13--17, 2026}{Dubai, United Arab Emirates}
\acmBooktitle{Proceedings of the ACM Web Conference 2026 (WWW '26), April 13--17, 2026, Dubai, United Arab Emirates}
\acmPrice{}
\acmDOI{10.1145/3774904.3792975}
\acmISBN{979-8-4007-2307-0/2026/04}
\newcommand{\shortname}{BalDRO}

\begin{document}

\title{BalDRO: A Distributionally Robust Optimization based Framework for Large Language Model Unlearning}

\author{Pengyang Shao}
\orcid{0000-0003-2838-1987}
\affiliation{%
  \institution{National University of Singapore}
  \country{Singapore}
}

\author{Naixin Zhai}
\orcid{0009-0005-3709-0163}
\affiliation{%
  \institution{University of Science and Technology of China}
  \city{Hefei}
  \country{China}
}

\author{Lei Chen}
\orcid{0000-0002-3193-7256}
\affiliation{%
  \institution{University of Science and Technology of China}
  \city{Hefei}
  \country{China}
}

\author{Yonghui Yang}
\orcid{0000-0002-7601-6004}
\affiliation{%
  \institution{National University of Singapore}
  \country{Singapore}
}

\author{Fengbin Zhu}
\authornotemark[1]
\email{zhfengbin@gmail.com}
\orcid{0000-0001-6776-2040}
\affiliation{%
  \institution{National University of Singapore}
  \country{Singapore}
}

\author{Xun Yang}
\authornote{Corresponding authors.}
\email{xyang21@ustc.edu.cn}
\orcid{0000-0003-0201-1638}
\affiliation{%
  \institution{University of Science and Technology of China}
  \city{Hefei}
  \country{China}
}

\author{Meng Wang}
\orcid{0000-0002-3094-7735}
\affiliation{%
  \institution{Hefei University of Technology}
    \city{Hefei}
  \country{China}
}

% \affiliation{%

% \affiliation{%

% \affiliation{%

% \affiliation{%

% \affiliation{%

% \affiliation{%

% \affiliation{%

% %%
% %% By default, the full list of authors will be used in the page
% %% headers. Often, this list is too long, and will overlap
% %% other information printed in the page headers. This command allows
% %% the author to define a more concise list
% %% of authors' names for this purpose.
\renewcommand{\shortauthors}{Pengyang Shao et al.}

\begin{abstract}
As Large Language Models (LLMs) increasingly shape online content, how to remove targeted information from well-trained LLMs (also known as LLM unlearning) has become increasingly critical for web governance. A key challenge in LLM unlearning lies in the sample-wise imbalance within the forget set: different samples exhibit widely varying unlearning difficulty, leading to asynchronous forgetting speeds where some knowledge remains insufficiently erased while others become over-forgotten. To address this challenge, we propose BalDRO, a novel and efficient framework for balanced LLM unlearning. BalDRO formulates unlearning as a min–sup process, where the inner process identifies a worst-case data distribution that adaptively emphasizes hard-to-unlearn samples, while the outer process updates model parameters based on the worst-case data distribution. We instantiate this formulation through two efficient variants: BalDRO-G, a discrete GroupDRO-based approximation that focuses on high-loss subsets, and BalDRO-DV, a continuous Donsker–Varadhan dual method that enables smooth, adaptive weighting within standard LLM training pipelines. Extensive experiments on the TOFU and MUSE benchmarks demonstrate the effectiveness of our proposed BalDRO, yielding significant improvements in both forgetting quality and model utility over existing methods. For reproducibility, we have released the code for BalDRO\footnote{\url{https://github.com/nxZhai/BalDRO}}. 
\end{abstract}





\maketitle

\section{Introduction}
As Large Language Models (LLMs) become increasingly embedded in web platforms and services~\cite{liu2025debate,yan2025agentsociety,bai2025learning,hu2024psycollm,xu2025multiagentesc}, ensuring that these models behave in a trustworthy and responsible manner has become essential for maintaining the reliability of web-based information ecosystems~\cite{qin2025explainable,xia2025identifying,zhou2025semi,zhou2025logic}.
A key aspect of achieving such reliability is the ability to remove outdated, incorrect, or privacy-sensitive knowledge from LLMs so that their behavior remains aligned with public values~\cite{geng2025comprehensive,ye2025towards,xu2025multi}, and safety requirements~\cite{ma2025attackseqbench,fang2025we}. 
This need is further reinforced by legal frameworks such as the General Data Protection Regulation (GDPR) and the California Consumer Privacy Act (CCPA)
, which mandate the “right to be forgotten’’ and require machine learning systems to support verifiable data erasure~\cite{zhao2024advanchor,zhao2024separable}. 
Together, these factors highlight the importance of developing effective LLM unlearning techniques~\cite{zhao2024makes,chen2025safeeraser}.

\begin{figure*}[t]
  \centering
  \includegraphics[width=0.85\textwidth]{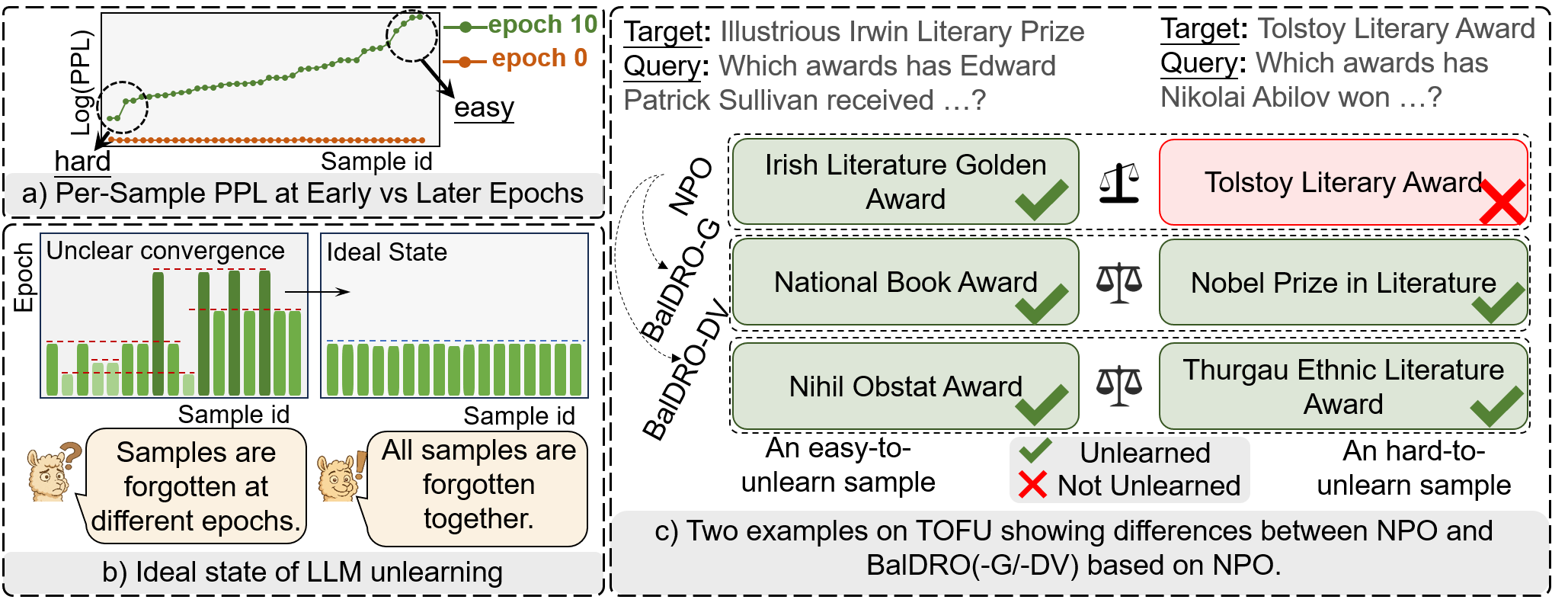}
  \caption{
    Illustration of sample-wise imbalance in LLM unlearning.
    a) Per-sample PPL (perplexity) at early and later epochs shows divergent forgetting dynamics, revealing heterogeneous unlearning difficulty in the forget set.
    b) This heterogeneity results in asynchronous convergence, whereas balanced unlearning aims to align forget epochs across samples. 
    c) Two real examples from the TOFU benchmark: for the easy sample, NPO successfully unlearns the target, whereas for the hard sample, NPO fails. In contrast, both BalDRO-G and BalDRO-DV successfully unlearn both cases.
    }
  \label{fig:examples}
\end{figure*}

Among the major challenges in achieving effective LLM unlearning, a particularly critical one lies in the {heterogeneous unlearning difficulty} across samples in the forget set~\cite{krishnan2025not,fan2024simplicity,zhangnegative}. 
Similar sample-level difficulty heterogeneity has also been observed in conditional generative models (e.g., progressive diffusion and unified conditional person generation frameworks) also exhibit strong heterogeneity across conditioning patterns, leading to uneven optimization difficulty~\cite{shen2024advancing,shen2024imagpose}. 
As illustrated in Figure~\ref{fig:examples}(a), when applying NPO~\cite{zhangnegative} to the same forget set, samples that start from similar initial states quickly diverge in their perplexity (PPL) trajectories at the same training epoch. 
This divergence reveals that different samples are forgotten at substantially different rates—some being easy to erase, while others remain resistant.
Figure~\ref{fig:examples}(b) further highlights this imbalance: each sample reaches its convergence point at a different epoch, making it difficult to determine when the entire forget set has been properly unlearned. 
This asynchronous forgetting dynamic is particularly problematic for gradient-based LLM unlearning methods, as such variants of the negative cross-entropy loss lack a well-defined upper bound~\cite{liu2025rethinking}. Continuing optimization to ensure full erasure of hard samples inevitably over-forgets the easy ones, thereby degrading overall model utility.
This phenomenon motivates a central question:  
\textbf{how can we achieve a balanced state where all forget samples are unlearned simultaneously?}

To solve the above question, recent studies have proposed various balanced unlearning paradigms~\cite{wang2024llm,yangexploring,krishnan2025not}. 
Krishnan et al. further show that the frequency of a fact in the pre-training corpus strongly influences its forgetting difficulty, and propose weighting schemes based on such frequency estimates, while pre-training corpus is often unavailable in practice~\cite{krishnan2025not}. 
Another line of balanced unlearning relies on reweighting each sample based on predefined or heuristic schemes, e.g., Negative Preference Optimization (NPO) leverages a reference model to stabilize gradient across different samples~\cite{zhangnegative}. Based on NPO, SimNPO removes the reference dependency with a uniform distribution~\cite{fan2024simplicity}.
A more advanced example in this reweighting family is SatImp, which assigns each sample a dynamic weight based on saturation and importance criteria~\cite{yangexploring}.
Despite their effectiveness, these approaches share a fundamental limitation—they rely on predefined or heuristic schemes that cannot dynamically adapt to the intrinsic distributional heterogeneity of data samples. 

To dynamically address the imbalance issue in LLM unlearning, we propose \shortname, a novel and efficient framework based on Distributionally Robust Optimization (DRO) that adaptively balances samples in the forget set. 
\shortname~ formulates LLM unlearning as a min–sup process: the outer process updates model parameters, while the inner process searches for a worst-case forget distribution within a KL-divergence uncertainty set. 
This adversarial distribution naturally assigns greater influence to samples with larger forget losses, thereby preventing the optimization from prematurely over-forgetting the easy samples.
To make this min–sup process tractable for LLMs, we provide two efficient realizations for the inner process. 
First, a GroupDRO-based variant (\shortname-G) approximates the adversarial distribution by dynamically selecting the highest-loss samples in each mini-batch, enforcing progress on the most under-forgotten regions of the data space.
Second, a Donsker–Varadhan dual formulation (\shortname-DV) offers a continuous alternative by converting the inner supremum into a smooth log-sum-exp goal, allowing \shortname~ to be seamlessly integrated into standard gradient-based unlearning pipelines without modifying model architectures or training loops.
Together, these realizations provide discrete and continuous views of the same underlying DRO principle—ensuring that forgetting progresses in a coordinated, balanced manner across all samples.  
As illustrated in Figure~\ref{fig:examples}(c), both variants based on NPO effectively unlearn samples of varying difficulty, achieving synchronized forgetting across easy and hard cases. 
Extensive experiments on the TOFU and MUSE benchmarks demonstrate that \shortname~ consistently achieves more synchronized forgetting dynamics, leading to higher forget quality while preserving general model utility more effectively than existing unlearning methods. 
Our contributions can be summarized as follows: 
\begin{itemize}[leftmargin=2em]
\item We systematically analyze the failure modes of existing gradient-based LLM unlearning methods and show that uncontrolled sample-wise forgetting imbalance is the key challenge to effective LLM unlearning. 
\item We propose \shortname, a novel and efficient DRO-based framework that balances samples.
Specifically, \shortname~ formulates LLM unlearning as a min–sup bi-level process, and provides two tractable realizations for the inner process. 
\item We conduct extensive experiments on TOFU and MUSE benchmarks, demonstrating the effectiveness of \shortname~ on achieving a better tradeoff between forgetting quality and model utility, e.g., on the TOFU benchmark, \shortname-G and \shortname-DV based on NPO improve forget quality by more than 20\% over NPO, while also delivering modest gains in model utility.
\end{itemize}

\section{Related Work}
\subsection{LLM Unlearning}
LLM unlearning aims to suppress knowledge contained in the forget set while preserving performance on the retain set~\cite{liu2025rethinking,jia2024soul,jeung2025dusk,liu2024large,zhai2026maximizing}, encompassing both model-editing–based approaches~\cite{shen2025llm,yu2025unierase,pan2025precise,li2025editing,pan2024finding} and gradient-based approaches~\cite{mekala2025alternate,fan2024simplicity,yangexploring}. 
Among them, we focus on gradient-based methods, as they are model-agnostic and compatible with LLM finetuning pipelines. Gradient-based methods can be broadly grouped into two categories. 
The first line of work is targeted unlearning, which defines explicit substitute outputs—typically refusal-style responses—for each forget-set query. These methods treat refusals as positive examples and enforce them using preference-based objectives such as DPO~\cite{rafailov2023direct}. 
Subsequent variants expand this paradigm: FLAT incorporates $f$-divergence–based loss adjustments~\cite{wangllm}, while AltPO generates diverse positive alternatives by substituting knowledge-relevant tokens~\cite{mekala2025alternate}.
However, targeted methods may induce shortcut behaviors~\cite{qisafety}, where the model learns to mimic refusal templates based on patterns in the input.

A second line of work, non-targeted unlearning, avoids constructing explicit target outputs and instead directly modifies gradients to reduce the influence of forget samples. 
This direction originates from gradient inversion and gradient correction methods such as GradAscent (GA) and GradDiff (GD). 
More principled formulations subsequently emerge: NPO~\cite{zhangnegative} casts unlearning as a negative log-likelihood objective; SimNPO~\cite{fan2024simplicity} removes the reference model with a uniform distribution; and SatImp~\cite{yangexploring} develops theoretically grounded criteria for loss reweighting.
Recent work further reveals that unlearning difficulty varies substantially across samples and correlates strongly with the frequency of knowledge occurrences in pretraining and finetuning~\cite{krishnan2025not}. 

\subsection{Distributionally Robust Optimization}
Distributionally Robust Optimization (DRO) provides a principled framework for learning models that remain reliable under distributional shifts or sampling uncertainty~\cite{lin2022distributionally}.
Instead of minimizing the expected loss over a single empirical distribution, DRO optimizes the worst-case loss within an uncertainty set of plausible distributions~\cite{yang2014distributed}, producing models that are demonstrably more tolerant to noise and variability in training data~\cite{cui2025learning,wang2024outlier}.

Building on this foundation, DRO has emerged as a central paradigm for enhancing robustness in a wide range of machine learning applications, particularly where distributional shifts or imbalanced sample difficulty are prevalent~\cite{lin2024temporally,liu2022distributionally}.
Recent studies further extend DRO to large language models (LLMs), showing that DRO-based objectives can effectively stabilize alignment under noisy or heterogeneous preference data~\cite{wutowards,zhu2025leveraging,xu2025robust}.
For instance, DRO-augmented preference optimization reduces the impact of pairwise and pointwise annotation noise in Direct Preference Optimization, leading to more reliable preference modeling~\cite{wutowards}.
However, {the use of DRO for LLM unlearning has not been explored}.
Given the inherent imbalance between hard-to-forget and easy-to-forget samples in unlearning,  DRO offers a natural and principled solution: its inner maximization automatically emphasizes worst-case (i.e., hardest) samples, aligning directly with the goal of balanced unlearning.
Therefore, in this paper, we focus on how to apply DRO to LLM unlearning.

\section{Preliminary}
\label{sec:pre}
In this section, we focus on analyzing several representative gradient-based LLM unlearning methods.
We start from a common formulation for gradient-based unlearning, typically expressed as a Gradient Difference (GD) objective: 
\begin{small}
\begin{equation}
\label{eq:basic_unlearn}
\ell_{\mathrm{all}}(\theta) \;=\; \ell_{f}(\theta) \;+\; \lambda \,\ell_{r}(\theta),
\end{equation}
\end{small}
where $\ell_f$ and $\ell_r$ denote the losses on the forget and retain sets, respectively. A classical choice for $\ell_f$ is  Gradient Ascent (GA), i.e., the reverse CE loss~\cite{yao2024large}: 

\begin{small}
\begin{equation}
\label{eq:ga_unlearn}
\mathcal{L}^{\mathrm{GA}}_f(\theta)
= \mathbb{E}_{(x,y)\sim D_f}
\big[ \log \pi_\theta(y \mid x) \big].
\end{equation}
\end{small}

Although GA effectively suppresses target likelihoods, it lacks any sample-wise stopping criterion~\cite{liu2025rethinking} and is highly sensitive to heterogeneous forgetting difficulty~\cite{krishnan2025not,yangexploring}, which leads to both under-forgotten and over-forgotten samples at convergence. 
To alleviate this issue, Negative Preference Optimization (NPO) introduces a reference model $\pi_{\mathrm{ref}}$ to balance samples~\cite{zhangnegative}:
\begin{small}
\begin{equation}
\label{eq:npo}
\ell^{\mathrm{NPO}}_f(\theta)
=
\mathbb{E}_{(x,y)\in {D}_f}\!\Big[
-\tfrac{2}{\alpha^{\mathrm{NPO}}}\,
\log \sigma\!\Big(
-\alpha^{\mathrm{NPO}} \log \tfrac{\pi_{\theta}(y\mid x)}{\pi_{\mathrm{ref}}(y\mid x)}
\Big)
\Big],
\end{equation}
\end{small}

where {\small$\alpha^{\mathrm{NPO}}$} is a temperature parameter that controls the sharpness of the penalty applied to the log-ratio between current policy $\pi_{\theta}(y\mid x)$ and the reference policy $\pi_{\mathrm{ref}}(y\mid x)$. 
Building on this idea, SimNPO removes the reference dependency and uses a length-normalized log-likelihood to obtain a simple, reference-free surrogate~\cite{fan2024simplicity}:
\begin{small}
\begin{equation}
\label{eq:simnpo}
\ell^{\mathrm{SimNPO}}_f(\theta)
=
\mathbb{E}_{(x,y)\in {D}_f}\!\left[
-\tfrac{2}{\alpha^{\mathrm{Sim}}}\,
\log \sigma\!\left(
-\tfrac{\alpha^{\mathrm{Sim}}}{|y|}\,\log \pi_{\theta}(y\mid x) 
\right)
\right].
\end{equation}
\end{small}

where $|y|$ denotes the output length used for normalization, and {\small$\alpha^{\mathrm{Sim}}$}  controls the sharpness of the logistic penalty for SimNPO.
Building upon these preference-based surrogate methods, SatImp further generalizes this line of work by introducing a {token-wise saturation-importance weight}, which adjusts the contribution of each generated token according to how confidently it is predicted.
Specifically, for the $k$-th token in $y$, the weight is defined as:
\begin{small}
\[
w^{\mathrm{SatImp}}_{x,y,k}
=
\pi_\theta\!\big(y_k \mid y_{<k},x\big)^{\alpha_1}
\big(1 - \pi_\theta\!\big(y_k \mid y_{<k},x\big)\big)^{\alpha_2},
\]
\end{small}
where the predicted probability $\pi_\theta(y_k \mid y_{<k}, x)$ determines both the {saturation} term and the {importance} term. 
The exponent $\alpha_1$ amplifies the effect of highly confident tokens, while $\alpha_2$ emphasizes low-probability tokens. The loss can be formulated as:
\begin{small}
\begin{equation}
\label{eq:satimp}
\ell^{\mathrm{SatImp}}_f(\theta)
=
\mathbb{E}_{(x,y)\in {D}_f}\!\left[
\sum_{k=1}^{|y|}
w^{\mathrm{SatImp}}_{x,y,k}\,
\log \pi_{\theta}\!\big(y_k \mid y_{<k},x\big)
\right].
\end{equation}
\end{small}

Although these weighting-based surrogate losses adaptively adjust sample contributions during training, they share a fundamental limitation: their updates are governed by pre-specified functional forms that dictate how weights respond to model predictions. These forms are heuristically designed rather than derived from a principled notion of balance, and thus cannot dynamically adapt to the evolving difficulty of different forget samples. As a result, existing methods often lead to asynchronous forgetting dynamics. In this regime, easy samples are quickly over-unlearned, while difficult ones remain insufficiently unlearned. 

\section{The Proposed Framework}
In this section, we start from presenting our proposed \shortname~ framework, including our unlearning goal based on a {principled notion of balance}, and corresponding bi-level process. 
Then, we propose two efficient and effective methods to realize the goal, which can be added as a plug-in to improve current LLM unlearning methods (e.g., NPO~\cite{zhangnegative}, SimNPO~\cite{fan2024simplicity}, and SatImp~\cite{yangexploring}). Finally, we discuss its cost as well as its connections to existing unlearning methods. 

\subsection{Overall Bi-level Process of \shortname}

To instantiate the principled notion of balance, we first formalize our goal as an optimization objective that explicitly captures sample-wise imbalance in the forget set.

\begin{definition}[Balanced Unlearning Objective]
\label{def:objective}
Given a forget dataset {\small${D}_f=\{z_i=(x_i,y_i)\}_{i=1}^n$} and the corresponding empirical distribution 
{\small$\widehat{D}_f(Z)=\tfrac{1}{n}\sum_i \delta_{z_i}(Z)$}, 
balanced LLM unlearning aims to find model parameters {\small$\theta$} that minimize the worst-case expected forget loss under distributional shifts:
\begin{small}
\begin{equation}
\label{eq:bal_dro_primal}
\arg\min_{\theta}\ 
\sup_{Q_f:\,\mathbb{D}(Q_f\Vert \widehat{D}_f)\le\eta}
\mathbb{E}_{Z \sim Q_f}\!\big[\ell_f(Z;\theta)\big].
\end{equation}
\end{small}
\end{definition}

Here, we apply DRO only to the forget loss, while the retain loss remains unchanged. Please note that, our experiments in Section \ref{sec:exp_ablation} show that extending DRO to the retain set brings negligible benefit, suggesting that retain samples already maintain a natural balance and do not require additional robustness adjustments.
{\small$Q_f$} denotes an adversarially perturbed distribution that reallocates probability mass toward harder forget samples, and {\small$\eta>0$} controls the uncertainty radius of the perturbation. 

Regarding the choice of divergence $\mathbb{D}$, any divergence that enables a valid min–sup DRO formulation could in principle be used; \shortname~is not restricted to a specific distance measure.
In this work, we adopt the Kullback–Leibler (KL) divergence~\cite{kullback1951kullback} because it provides a closed-form Donsker–Varadhan (DV) dual representation, which leads to a tractable and numerically stable implementation at LLM scale while naturally inducing an exponential reweighting aligned with forgetting difficulty: 
\begin{small}
\begin{equation}
\label{eq:KL}
\mathbb{D}_{\mathrm{KL}}(Q_f\Vert \widehat{D}_f)
= \mathbb{E}_{Z \sim Q_f}\!\left[\log\frac{Q_f(Z)}{\widehat{D}_f(Z)}\right] \le \eta.
\end{equation}
\end{small}

\begin{figure}[t]
  \centering
  \includegraphics[width=0.45\textwidth]{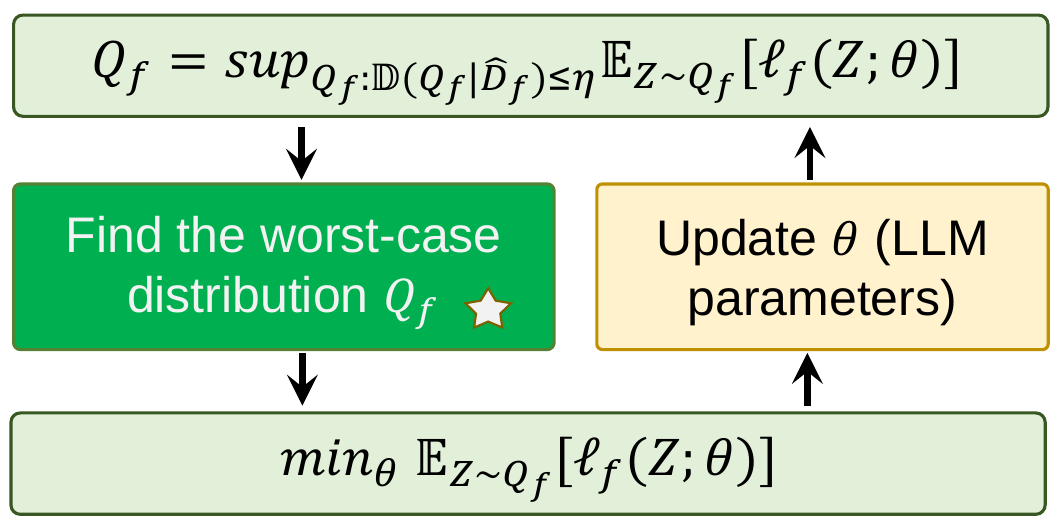}
  \caption{The overall {min--sup} process of \shortname. The inner ``sup'' adaptively identifies the hardest forget distribution, while the outer ``min'' optimizes model parameters.}
  \label{fig:model}
\end{figure}

As shown in Figure \ref{fig:model}, this bi-level process embodies our core objective: the inner ``sup'' identifies a worst-case forget distribution within the KL ball-thereby adaptively emphasizing hard-to-forget samples—while the outer ``min'' updates model parameters to minimize this adversarial loss.  
The resulting optimization naturally balances forgetting progress across samples without explicit heuristic weighting, grounding our method in a theoretically robust notion of equilibrium. 
Obviously, the outer process can be easily realized by finetuning LLM, and the key challenge is how to assess the inner process (i.e., $\sup_{Q_f:\,\mathbb{D}_{\mathrm{KL}}(Q_f\!\Vert \widehat{D}_f)\le\eta}\mathbb{E}_{Z\sim Q_f}\!\big[\ell_f(Z;\theta)\big]$), which captures the worst-case expected loss under distributional uncertainty 
around the empirical forget set. 
% Eq.~(\ref{eq:bal_dro_primal}) lies in evaluating the inner process, formulated as: 
% \begin{small}
% \begin{equation}
% \sup_{Q_f:\,\mathbb{D}_{\mathrm{KL}}(Q_f\!\Vert \widehat{D}_f)\le\eta}
% \mathbb{E}_{Z\sim Q_f}\!\big[\ell_f(Z;\theta)\big],
% \label{eq:inner}
% \end{equation}
% \end{small}
% which captures the worst-case expected loss under distributional uncertainty 
% around the empirical forget set.

\subsection{The Inner Process}
To instantiate Definition~\ref{def:objective}, we develop two tractable realizations of the inner process in Eq.~(\ref{eq:bal_dro_primal}). 
The first, {BalDRO-G}, provides a discrete approximation based on Group Distributionally Robust Optimization, 
while the second, {BalDRO-DV}, offers a continuous dual formulation grounded in the Donsker--Varadhan (DV) representation. 
Together, they provide complementary discrete and continuous perspectives of realizing our goal, which both follow the balanced unlearning principle.

\subsubsection{BalDRO-G: Discrete Group-Level Approximation}

Group Distributionally Robust Optimization (GroupDRO) offers a tractable approximation to the inner process in Definition~\ref{def:objective} by explicitly selecting the most difficult subsets within the forget set~\cite{zhang2023stochastic}. 
Concretely, we partition the empirical forget set {\small${D}_f$} into {\small$G$} groups
{\small$\{{D}_f^{(1)}, {D}_f^{(2)}, \dots, {D}_f^{(G)}\}$}, and optimize the maximum expected loss among them:
\begin{small}
\begin{equation}
 \max_{g=1,\dots,G}\;
\mathbb{E}_{z_i \sim {D}_f^{(g)}}\!\big[\ell_f(z_i;\theta)\big].
\end{equation}
\end{small}

Intuitively, this formulation serves as a discrete instantiation of the inner process in Definition~\ref{def:objective}: 
the adversarial distribution {\small$Q_f$} is approximated by assigning all probability mass to the group with the largest expected forget loss. 
This allows BalDRO-G to operationalize the theoretical {min--sup} objective through a finite, group-level approximation that is both stable and computationally efficient in practice. 
At each iteration, BalDRO-G enforces progress on the worst-performing group, directing the optimization toward the under-forgotten region of the data space. 
Rather than relying on predefined grouping schemes, we adopt a percentile-based strategy (top-50\%) to avoid the instability of extreme single-sample maxima while still focusing on the hardest region of the distribution.
%This design follows standard practice in robust optimization and is further validated in Appendix~\ref{sec:appendix-exp}. 

\subsubsection{BalDRO-DV: Continuous Dual Adaptive Weighting}
While GroupDRO captures coarse-grained group imbalance, it still depends on manually defined partitions.
To derive a fully continuous and differentiable relaxation of the same inner process, we revisit the original DRO problem and apply a Lagrangian relaxation~\cite{lemarechal2001lagrangian} to the KL-divergence constraint:
\begin{small}
\begin{equation}
\sup_{Q_f}\;
\Big\{
\mathbb{E}_{Z \sim Q_f}\!\big[\ell_f(Z;\theta)\big]
-
\beta
\big(
\mathbb{D}_{\mathrm{KL}}(Q_f \Vert \widehat{D}_f) - \eta
\big)
\Big\},
\label{eq:bal_dro_lagrangian}
\end{equation}
\end{small}
where \(\beta>0\) is a Lagrange multiplier.
The constant term \(\beta\eta\) is independent of \(Q_f\) and can be ignored during maximization.
Substituting the definition of the KL divergence yields:
\begin{small}
\[
\sup_{Q_f}\;
\mathbb{E}_{Z \sim Q_f}\!\left[
\ell_f(Z;\theta)
-
\beta \log\frac{Q_f(Z)}{\widehat{D}_f(Z)}
\right].
\]
\end{small}
Maximizing the inner expression with respect to \(Q_f\) gives the optimal adversarial distribution. The detailed derivation is provided in Appendix~\ref{sec:closed_form}. Here, we directly provide the closed-form solution as follows: 
\begin{small}
\begin{equation}
Q_f^\star(Z)
=
\frac{
\widehat{D}_f(Z)\,
\exp\!\big(\ell_f(Z;\theta)/\beta\big)
}{
\mathbb{E}_{Z' \sim \widehat{D}_f}\!\left[\exp\!\big(\ell_f(Z';\theta)/\beta\big)\right]
}.
\label{eq:bal_dro_qstar}
\end{equation}
\end{small}
This distribution exponentially upweights samples (or regions of the data space) 
with higher forget loss, thereby implementing an adaptive, distribution-level reweighting mechanism. Substituting {\small\(Q_f^\star\)} back into Eq.~\eqref{eq:bal_dro_primal} yields a closed-form expression for the inner supremum:
\begin{small}
\begin{equation}
\min_{\theta}\ 
\beta \eta
+
\beta
\log
\mathbb{E}_{Z \sim \widehat{D}_f}
\!\left[
\exp\!\left(\frac{\ell_f(Z;\theta)}{\beta}\right)
\right],
\label{eq:bal_dro_dv}
\end{equation}
\end{small}
which corresponds to the continuous Donsker–Varadhan (DV) representation. Although $\beta$ can in principle be optimized jointly, we adopt a fixed-$\beta$ variant for better stability and computational simplicity~\cite{wutowards}. Finally, minimizing Eq.~\eqref{eq:bal_dro_dv} leads to the {DV dual formulation}:
\begin{small}
\begin{equation}
\begin{aligned}
\min_{\theta}\;
\beta
\log
\mathbb{E}_{Z \sim \widehat{D}_f}
\!\left[
\exp\!\left(\frac{\ell_f(Z;\theta)}{\beta}\right)
\right]  = \min_{\theta}\; \beta
\log
\left(
\frac{1}{n}\sum_{i=1}^{n}
\exp\!\left(\frac{\ell_f(z_i;\theta)}{\beta}\right)
\right).
\label{eq:bal_dro_final}
\end{aligned}
\end{equation}
\end{small}

The equality in Eq.~\eqref{eq:bal_dro_final} follows from the definition of the empirical distribution {\small$\widehat{D}_f(Z)=\frac{1}{n}\sum_{i=1}^{n}\delta_{z_i}(Z)$},
under which the expectation {\small$\mathbb{E}_{Z\sim\widehat{D}_f}[\cdot]$} 
is equivalent to calculating over all samples in the forget set.
This dual formulation transforms the intractable distributional optimization problem into a smooth and differentiable {log-sum-exp} goal, 
which can be seamlessly integrated into existing LLM unlearning pipelines.
Intuitively, the exponential term {\small$\exp(\ell_f(z_i;\theta)/\beta)$} adaptively scales gradient magnitudes: harder samples receive higher weighting, while already-forgotten ones are naturally downweighted, achieving a self-regulating balance across the forget set. 
This adaptive weighting mechanism serves as a continuous, theoretically grounded realization  of the balanced unlearning principle introduced in Definition~\ref{def:objective}.

\subsection{Model Discussion}

\subsubsection{Additional Time Complexity}
BalDRO introduces only marginal computational overhead compared to standard unlearning objectives. 
For {BalDRO-G}, given a batch of $n$ forget samples, we first compute their per-sample forget losses {\small$\{\ell_f\}_{i=1}^{n}$} 
and identify the hardest subset (e.g., top-$50\%$) for back-propagation. 
This selection can be implemented via a partial sort with at most {\small$\mathcal{O}(n \log n)$} complexity, 
which is negligible relative to the cost of LLM forward and backward passes.

For {BalDRO-DV}, the only additional cost arises from the log-sum-exp term in Eq.~\eqref{eq:bal_dro_final}, 
which requires computing one exponential and one logarithm per sample within each mini-batch. 
Specifically, it involves three lightweight element-wise operations: 
scaling by {\small$1/\beta$}, evaluating {\small$\exp(\ell_i/\beta)$}, and performing a batch-level reduction via 
{\small$\log(\sum_i \exp(\ell_i/\beta))$}. 
These operations result in linear complexity {\small$\mathcal{O}(n)$}, the same order as the base loss computation, 
with only a small constant factor overhead due to exponentiation and normalization.

% In summary, both BalDRO-G and BalDRO-DV preserve the scalability of existing unlearning pipelines, 
% adding only minimal constant-time operations while maintaining the same asymptotic complexity. 

\subsubsection{Relations to Existing LLM Unlearning Methods}
BalDRO is designed as a plugin-style framework for gradient-based LLM unlearning, offering strong generality across a wide range of existing methods. This design choice is intentional: it reveals that distributional balancing is a missing yet broadly beneficial component across diverse unlearning objectives. Rather than tailoring to any specific loss, BalDRO can capture a fundamental property of unlearning dynamics, which will be evaluated in next section. 

Moreover, BalDRO provides a unified perspective that encompasses existing unlearning approaches as special cases under different regimes of the temperature parameter~$\beta$. In the discrete variant (BalDRO-G), selecting the hardest subset corresponds to the limiting case of optimizing the maximum loss over grouped samples; and as the subset size increases, the behavior naturally approaches that of the original unlearning method. In the continuous variant (BalDRO-DV), the formulation in Eq.~\eqref{eq:bal_dro_final} generalizes this trajectory: as $\beta \to \infty$, the exponential weighting becomes uniform and the objective degenerates to the standard mean-loss formulation (i.e., no balancing), whereas as $\beta \to 0$, the log-sum-exp term approaches the maximum loss, recovering the worst-case optimization equivalent to BalDRO-G when each sample is treated as its own group.

\section{Experiments}
In this section, we try to answer these research questions: 
\begin{itemize}[leftmargin=1em]
    \item RQ 1: Does \shortname~ provide stable improvements on different gradient-based LLM unlearning objectives? (Sec.~\ref{sec:overall_perf})

    \item RQ 2: Does \shortname~ perform well across varying sizes of the forget set? (Sec.~\ref{sec:exp_varying_size})
        
    \item RQ 3: Is \shortname~ robust to hyperparameter settings? (Sec.~\ref{sec:exp_hyperparameters})

    \item RQ 4: Is \shortname~ also effective on the retain set? (Sec.~\ref{sec:exp_ablation})
    
    \item RQ 5: Is \shortname~ still effective on different metrics? (Sec.~\ref{sec:other_metrics})
\end{itemize}
% Due to page limits, we have placed additional experiments in Appendix \ref{sec:appendix-exp}, including varying the sample selection ratio in \shortname-G. 

\begin{table*}[ht]
    \centering
    \renewcommand{\arraystretch}{1.1}
    \resizebox{0.85\textwidth}{!}{
        \begin{tabular}{c|ccccccccc}
        \hline
\textbf{Method} & \textbf{FQ (↑)} & \textbf{MU (↑)} & \textbf{Fluency (↑)} & \textbf{EM (↓)} & \textbf{ES (↓)} & \textbf{F-TR (↑)} & \textbf{Ra-TR (↑)} & \textbf{R-TR (↑)} & \textbf{Rw-TR (↑)} \\ \hline
        Original          & 0.0013 & 0.6276 & 0.8889 & 1.0000 & 1.0000 & 0.5306 & 0.6120 & 0.4596 & 0.5521 \\
        Retain          & 1.0000 & 0.6268 & 0.9222 & 0.7121 & 0.0719 & 0.6777 & 0.6087 & 0.4621 & 0.5612 \\ \hline
        GradAscent & 0.2657 & 0.5313 & 0.5326 & 0.5572 & 0.0361 & 0.0019 & 0.6049 & 0.4357 & 0.5848 \\
        GradDiff & 0.5786 & 0.6064 & 0.3976 & 0.5397 & 0.0481 & 0.6359 & 0.5595 & 0.4523 & 0.5720 \\ \hdashline
        % DPO & 0.4046 & 0.5547 & 0.9330 & 0.8026 & 0.1281 & 0.7342 & 0.6114 & 0.3932 & 0.5437 \\ \hdashline
        NPO & 0.7659 & 0.5775 & 0.8158 & {0.6842} & 0.0982 & {0.7125} & 0.5807 & {0.4360} & 0.5489 \\ \rowcolor[gray]{0.9}
        \textit{NPO + BalDRO-G}  & {0.9188} & \textbf{0.6126} & {0.8220} & 0.7634 & {0.0630} & 0.6859 & \textbf{0.6187} & \textbf{0.4495} & \textbf{0.5741} \\ \rowcolor[gray]{0.9}
        \textit{NPO + BalDRO-DV} & \textbf{0.9900} & {0.5815} & \textbf{0.8227} & \textbf{0.6659} & \textbf{0.0593} & \textbf{0.7238} & {0.6000} & 0.4148 & {0.5610} \\ \hdashline
        SimNPO & 0.4046 & 0.5643 & 0.8422 & 0.7383 & 0.0850 & 0.6768 & {0.6025} & 0.3990 & 0.5377 \\ \rowcolor[gray]{0.9}
        \textit{SimNPO + BalDRO-G} & \textbf{0.5786} & {0.5651} & \textbf{0.8894} & {0.7189} & {0.0633} & \textit{0.7034} & 0.5877 & \textbf{0.4307} & {0.5712} \\ \rowcolor[gray]{0.9}
        \textit{SimNPO + BalDRO-DV}& \textbf{0.5786} & \textbf{0.5917} & {0.8479} & \textbf{0.6926} & \textbf{0.0521} & \textbf{0.7257} & \textbf{0.6345} & {0.4215} & \textbf{0.5726} \\ \hdashline
        SatImp & 0.0013 & 0.5342 & {0.8272} & 0.9466 & {0.2041} & 0.5117 & 0.5493 & 0.4315 & {0.5338} \\ \rowcolor[gray]{0.9}
        \textit{SatImp + BalDRO-G}  & \textbf{0.0971} & \textbf{0.6003} & \textbf{0.8520} & {0.8879} & \textbf{0.1609} & \textbf{0.6108} & \textbf{0.6131} & \textbf{0.4654} & \textbf{0.5480} \\ \rowcolor[gray]{0.9}
        \textit{SatImp + BalDRO-DV}     & {0.0068} & {0.5480} & 0.7588 & \textbf{0.8646} & 0.4522 & {0.5515} & {0.5764} & {0.4485} & 0.4973 \\ \hline
        \end{tabular}
    }
    \caption{Overall performance on TOFU benchmark with forget ratio = 1\%. We bold the best result.}
    \label{tab:tofu_main_table}
\end{table*}
\begin{table*}[ht]
\centering
\renewcommand{\arraystretch}{1.1}
\resizebox{0.85\textwidth}{!}{
\begin{tabular}{c|cccc|cccc}
\hline
\multirow{2}{*}{\textbf{Method}} &
\multicolumn{4}{c|}{\textbf{News}} &
\multicolumn{4}{c}{\textbf{Books}} \\ \cline{2-9}
& \textbf{KM-Dr} (↑) & \textbf{KM-Df} (↓) & \textbf{VM-Df} (↓) & \textbf{PL} ($\rightarrow$ 0)
& \textbf{KM-Dr} (↑) & \textbf{KM-Df} (↓) & \textbf{VM-Df} (↓) & \textbf{PL} ($\rightarrow$ 0) \\ 
\hline
Original & 0.5552 & 0.6443 & 0.5789 & -99.8111
         & 0.6913 & 0.4712 & 0.9970 & -57.3410 \\
Retain   & 0.5602 & 0.3279 & 0.2016 & -4.7200
         & 0.6874 & 0.3029 & 0.1445 & 8.1600 \\ \hline
GradAscent & 0.0000 & 0.0000 & 0.0000 & 25.1259
           & 0.0000 & 0.0000 & 0.0000 & -22.8180 \\
GradDiff & 0.2519 & 0.2938 & 0.0029 & 108.9840
         & 0.0678 & 0.0089 & 0.0035 & -37.0562 \\ \hdashline
NPO & 0.4552 & 0.5978 & 0.4255 & -90.8480
    & 0.6424 & 0.4414 & 0.6011 & -55.7692 \\
\rowcolor[gray]{0.9}
\textit{NPO+BalDRO-G} & \textbf{0.4626} & {0.5805} & \textbf{0.3826} & \textbf{-65.7011}
                      & {0.6486} & {0.4376} & {0.5465} & \textbf{-54.2160} \\
\rowcolor[gray]{0.9}
\textit{NPO+BalDRO-DV} & {0.4589} & \textbf{0.5754} & {0.3934} & {-69.5214}
                       & \textbf{0.6520} & \textbf{0.4107} & \textbf{0.5230} & {-55.2515} \\ \hdashline
SimNPO & 0.4121 & {0.5806} & 0.3829 & {-99.8951}
       & {0.5969} & 0.3009 & 0.2364 & -51.7018 \\
\rowcolor[gray]{0.9}
\textit{SimNPO+BalDRO-G} & {0.4272} & 0.5940 & 0.4193 & \textbf{-99.8950}
                         & \textbf{0.6151} & \textbf{0.2841} & {0.2264} & {-51.2944} \\
\rowcolor[gray]{0.9}
\textit{SimNPO+BalDRO-DV} & \textbf{0.4571} & \textbf{0.5670} & \textbf{0.1829} & 100.4139
                          & 0.5393 & 0.3009 & \textbf{0.1935} & \textbf{-49.3898} \\ \hdashline
SatImp & 0.3797 % 0.4097
& 0.5902 & 0.4403 & -99.8951
       & {0.6026} % tips: here pay attention 0.6202
       & 0.4017 & 0.8730 & -58.3395 \\
\rowcolor[gray]{0.9}
\textit{SatImp+BalDRO-G} & {0.3805} & {0.5053} & {0.4197} & {-99.8531}
                         & \textbf{0.6037} & {0.3802} & \textbf{0.4955} & \textbf{-54.5858} \\
\rowcolor[gray]{0.9}
\textit{SatImp+BalDRO-DV} & \textbf{0.3967} & \textbf{0.4568} & \textbf{0.3552} & \textbf{-99.8321}
                          & 0.6013 & \textbf{0.3672} & {0.5535} & {-57.2485} \\ \hline
\end{tabular}
}
\caption{Overall performance on the MUSE benchmark. MUSE adopts a fixed forget/retain split, and we bold the best result.}
\label{tab:muse_news_books_combined}
  \vspace{-1.0em}
\end{table*}

\subsection{Experiment Setup}

\subsubsection{Benchmarks} 
We evaluate \shortname\ on two complementary unlearning benchmarks that together capture both controlled and realistic forgetting behaviors. 
{TOFU}~\cite{maini2024tofu} contains synthetic QA pairs for 200 fictional authors, ensuring that all forget-set knowledge originates solely from fine-tuning. 
{TOFU} includes three unlearning settings, aiming to remove 1\%, 5\% and 10\% of the total dataset.
This setup isolates unlearning effects from pretraining priors, providing a controlled environment for analyzing forgetting dynamics.  
{MUSE}~\cite{shi2024muse} consists of real-world books and news articles, where forget and retain splits are semantically entangled. 
It thus presents a more realistic and challenging testbed for assessing the robustness and practicality of unlearning algorithms.

\subsubsection{Evaluation Metrics}
{TOFU} offers fine-grained controls for separately assessing forgetting and retention.
Forget Quality (FQ) captures alignment with a retain-only reference using the KS distance over Truth Ratio distributions, whereas Model Utility (MU) evaluates the preservation of non-forget knowledge via the harmonic mean of Probability, ROUGE, and Truth Ratio on $D_r$ and other holdout sets.
Extraction Memorization (EM) and Extraction Strength (ES) quantify verbatim memorization at the token and prefix levels.
We additionally report four Truth Ratio variants—F-TR, Ra-TR, R-TR, and Rw-TR—which together assess forgetting quality, robustness to unseen real authors, retention correctness, and real-world consistency.
Lower EM and ES indicate stronger forgetting, while higher FQ, MU, and Truth Ratio values reflect better retention and balanced unlearning. 

{MUSE} contains real-world texts, such as novels and news articles, where the forget and retain sets exhibit strong semantic overlap, making targeted unlearning substantially more challenging.
To evaluate both semantic and lexical forgetting as well as privacy risks, we adopt three metrics: KnowMem (KM), VerbMem (VM), and PrivLeak (PL).
KM measures semantic recall, and VM assesses verbatim recall through exact lexical overlap; both are computed on the forget set (KM-Df, VM-Df) and the retain set (KM-Dr) to distinguish desired forgetting from unintended over-forgetting.
PL estimates membership-inference risk, indicating whether forgotten samples remain distinguishable from unseen data.
Lower KM-Df and VM-Df indicate stronger forgetting and privacy protection, while higher KM-Dr reflects better retention.
For PL, values closer to zero are preferred.

\subsubsection{Baseline Methods.} 
We compare our method with several baselines. 
Original refers to the model trained on both forget and retain sets; it represents the starting point of unlearning. 
In contrast, Retain denotes the model fine-tuned only on the retain set, reflecting an idealized upper bound for LLM unlearning. 
The remaining baselines, introduced in Section~\ref{sec:pre}, cover the classic and recent gradient-based unlearning methods. 
Specifically, we include GradAscent (GA)~\cite{yao2024large}, which performs gradient ascent on forget samples to reduce the model’s confidence, and GradDiff (GD)~\cite{yao2024large}, which additionally leverages the retain set to mitigate unnecessary utility loss. 
Moreover, we evaluate \shortname~on three representative and state-of-the-art unlearning frameworks—NPO~\cite{zhangnegative}, SimNPO~\cite{fan2024simplicity}, and SatImp~\cite{yangexploring}. 
These baselines constitute the latest gradient-based unlearning techniques, while editing-based approaches are not included~\cite{shen2025llm,yu2025unierase,zeng2025visual}, as they follow a fundamentally different model-editing paradigm that is incompatible with us.

\subsubsection{Implementation Details.}
All experiments are conducted on a server equipped with 8 NVIDIA A800 GPUs.
We use LLaMA-2-7B~\cite{touvron2023llama} as the primary backbone model for evaluating overall unlearning performance.
For all methods, we perform a  hyperparameter search over learning rates in $\{1 \times 10^{-5}, 2 \times 10^{-5}, 5 \times 10^{-5}, 1 \times 10^{-4}\}$, batch sizes in $\{8, 16, 32\}$, $\beta$ in $\{1.0, 2.0, 5.0, 10.0\}$ and $\lambda$ in $\{0.25, 0.5, 1.0, 2.0\}$.
%This setup ensures a fair and consistent comparison across methods, allowing the reported performance to reflect the best achievable results under each configuration. 

\begin{figure*}[t]
  \centering
  \includegraphics[width=\textwidth]{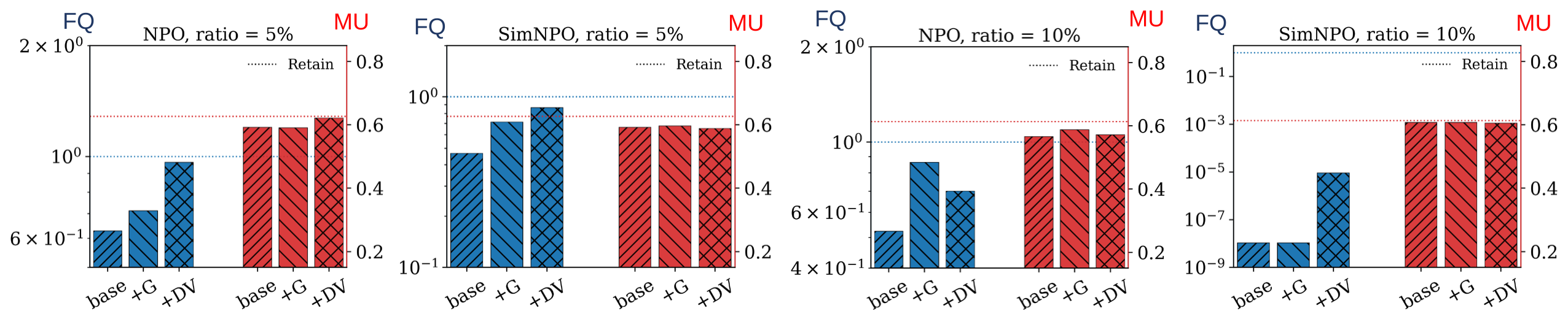}
  \caption{
Performance with varying forget ratios (5\% and 10\%) on the TOFU benchmark.
We focus on FQ and MU, the two most commonly used metrics on TOFU, and select NPO and SimNPO as base methods due to their strong overall performance on these two metrics. 
Here, “+G” denotes our proposed \shortname-G, and “+DV” corresponds to \shortname-DV. 
}
  \label{fig:tofu_results_forget05_10}
\end{figure*}
\subsection{Overall Performance~(RQ1)}
\label{sec:overall_perf}

\subsubsection{Main Results on TOFU}

Table~\ref{tab:tofu_main_table} presents the results under the TOFU benchmark. We summarize three observations. 
First, \shortname~variants consistently improve FQ and reduce EM/ES across most base methods. 
Second, different base methods benefit to different degrees. SimNPO obtains the most stable overall gains: FQ rises from 0.4046 to 0.5786, EM decreases, and several TR metrics also improve. 
In contrast, SatImp shows the largest relative jump in FQ (0.0013→0.0971 (\shortname-G), over \textbf{70×}), mainly because its baseline forgetting is extremely weak, leaving enough space for improvements. 
Third, \shortname-DV and \shortname-G show different strengths across backbones. 
For NPO and SimNPO, \shortname-DV typically performs better, achieving the highest FQ and lower EM/ES while maintaining competitive TR performance. 
For SatImp, however, \shortname-G works better, giving both the strongest FQ improvement and more stable TR results.

\subsubsection{Main Results on MUSE}
Table~\ref{tab:muse_news_books_combined} summarizes the results on the MUSE benchmark. We have several observations from this table. 
First, we observe that both \shortname-G and \shortname-DV consistently enhance the base methods (NPO, SimNPO, SatImp) across domains. This demonstrates the general effectiveness of our framework: \shortname~ improves forget quality while maintaining LLM utility. For instance, on the News domain, \shortname-G based on NPO lowers VM-Df (0.4255→0.3826) and improves PL (–90.85→–65.70).
Second, across both domains, \shortname-G and \shortname-DV perform comparably.  Rather than one variant uniformly dominating the other, they exhibit complementary strengths across settings. 
Given that \shortname-DV only improves the objective function and is therefore simpler to implement, it may be the more practical choice in real-world deployments.
Finally, the three base methods (NPO, SimNPO, SatImp) respond to \shortname~ with different degrees of improvement. NPO already delivers strong baseline, limiting the potential gains. E.g., in the Books domain, \shortname-G yields modest changes in KM-Dr (0.6424→0.6486) and PL (–55.77→–54.22). 
In contrast, SimNPO and SatImp perform worse, giving \shortname~ substantially more room to provide larger improvements across different metrics.

\subsection{Detailed Model Analyses}
\label{sec:analysis}

\subsubsection{Varying Forget Ratios (RQ 2)}
\label{sec:exp_varying_size}
% 为了验证 BalDRO 在不同规模的遗忘集合上是否能够保持稳定效果，我们在 TOFU 基准上进一步进行了 forget ratio = 5% 与 10% 的额外实验。Fig.~\ref{fig:tofu_results_forget05_10} 展示了基于 NPO 与 SimNPO 的全部结果，我们从中得到如下观察。
% 首先，在两种 base methods（NPO / SimNPO）下，BalDRO-G 与 BalDRO-DV 在保持与原方法可比的 MU 前提下，都取得了显著的 FQ 改善。
% 以 forget ratio = 5% 为例，NPO 的 FQ 从0.6284提升至0.7125（+G）与0.9646（+DV）；SimNPO 的 FQ 则从0.4662进一步提升到接近0.7125（+G）与 0.8655（+DV）。这些结果表明，BalDRO 在不同规模的遗忘集合上均具备一致且显著的提升能力。
% 其次，我们观察到 BalDRO-DV 在绝大多数情况下的 FQ 上都优于 BalDRO-G。
% 在 NPO（ratio = 5%）与 SimNPO（ratio = 5%）上，DV 均比 G 额外带来10%以上的提升；在更难的 NPO（ratio = 10%）与 SimNPO（ratio = 10%）设置中，DV 的增益更为稳定，而 BalDRO-G 的提升幅度相对有限。这表明，在实际部署时，BalDRO-DV 更适合作为默认选择。
% 最后，我们注意到 BalDRO 在不同 forget ratios 下对 MU 的影响一直保持在非常小的范围内。
% 无论是基于 NPO 还是 SimNPO、无论 forget ratio = 5% 还是 10%，MU 基本都保持在 retain 水平附近，说明 BalDRO 的忘却行为是“精准且不伤整体能力”的，不会破坏模型对非遗忘样本的整体表现。
To verify whether our proposed \shortname~ maintains stable performance across different sizes of the forget set, we conduct additional experiments on TOFU using forget-set ratios of 5\% and 10\%.

Figure \ref{fig:tofu_results_forget05_10} presents the complete results based on two representative methods (NPO and SimNPO), from which we make the following observations. First, under both two methods, \shortname-G and \shortname-DV consistently achieve substantial improvements in FQ while maintaining MU comparable to the original models. 
For instance, with a forget ratio of 5\%, the FQ of NPO increases from 0.6284 to 0.7125 (\shortname-G) and 0.9646 (\shortname-DV); for SimNPO, FQ improves from 0.4662 to 0.7125 (\shortname-G) and 0.8655 (\shortname-DV). 
These results indicate that \shortname~ provides consistent and significant gains regardless of the size of the forget set. 
Second, we observe that \shortname-DV outperforms \shortname-G in FQ in most cases.
For ratio = 5\%, \shortname-DV delivers more than 10\% additional improvement over G; and under the more challenging settings of ratio = 10\%, \shortname-DV’s gains remain stable, while the improvements from \shortname-G are relatively limited. 
This suggests that \shortname-DV is the preferable choice for practical deployment due to its more reliable performance.
Finally, \shortname’s impact on MU remains minimal regardless of whether the forget ratio is 5\% or 10\%, indicating that \shortname~ does not harm general ability. 

\begin{figure}[t]
  \centering
  \includegraphics[width=\columnwidth]{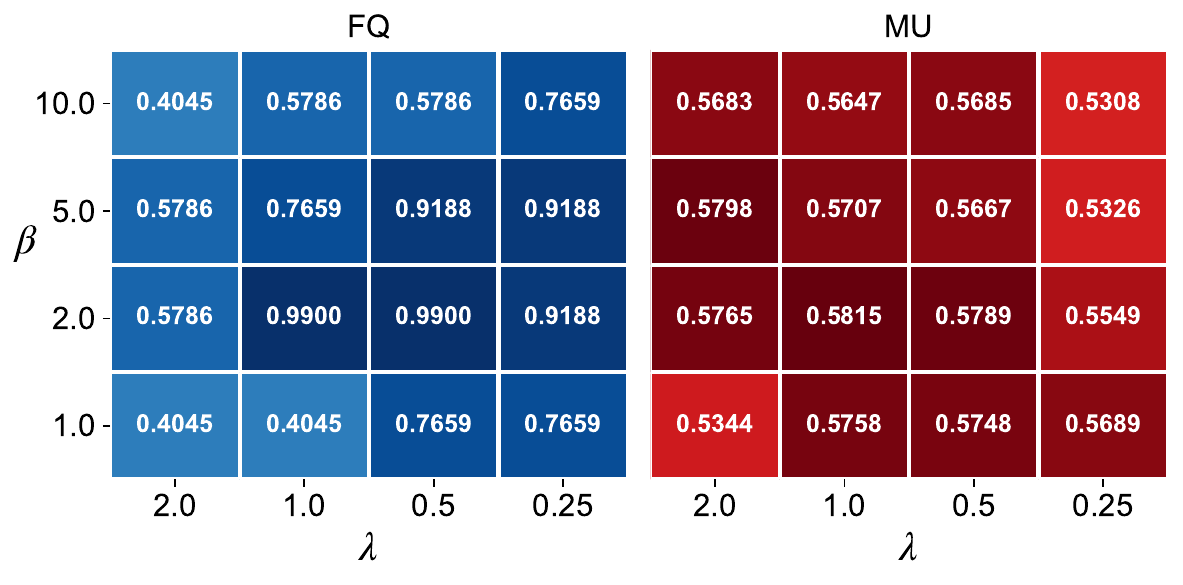}
  \caption{Performance of \shortname-DV with varying $\beta$ and balancing parameter $\lambda$ on the TOFU benchmark.
}
  \label{fig:beta}
\end{figure}

\begin{figure}[t]
  \centering
  \includegraphics[width=\columnwidth]{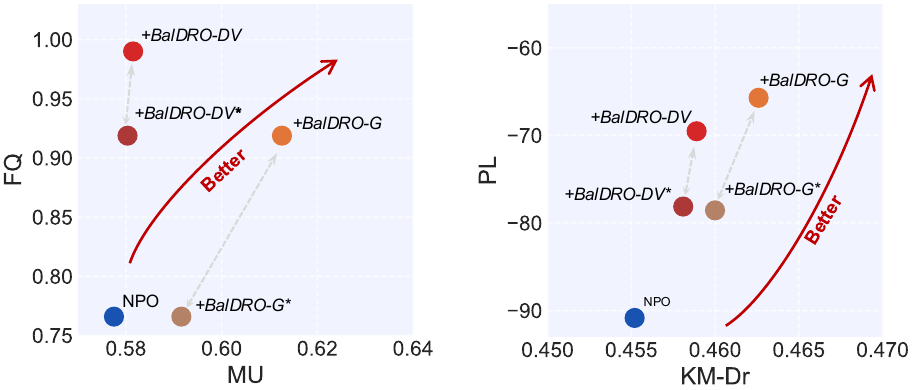}
  \caption{Performance comparisons between whether applying DRO to the retain loss $\ell_{r}(\theta)$ on TOFU and MUSE benchmarks.  `*' indicates DRO is applied to the retain loss.}
  \label{fig:retain-DRO}
  \vspace{-1.0em}
\end{figure}

\subsubsection{Hyperparameter Analyses (RQ 3)}
\label{sec:exp_hyperparameters}
Figure~\ref{fig:beta} illustrates the influence of the coefficient $\beta$ in Eq.~(\ref{eq:bal_dro_final}) and the balancing parameter $\lambda$ in Eq.~(\ref{eq:basic_unlearn}) on unlearning performance in the TOFU benchmark.
First, FQ peak at $\beta = 2.0$ and $\lambda = 1.0$, which suggests the presence of a well-defined optimal region where the adversarial reweighting and the base unlearning objective reinforce each other most effectively. 
MU remains relatively stable across configurations, showing only slight degradation when parameters move toward extreme values, indicating that the model is generally robust to moderate variations in both hyperparameters.
Second, we observe that $\beta = 2$ or $5$ produces the most reliable and consistent improvements. 
As discussed earlier, excessively large $\beta$ dilutes the effect of sample-wise weighting and causes the DRO objective to collapse toward a near-uniform distribution, thereby reducing its ability to correct imbalance. 
Conversely, overly small $\beta$ forces the model to concentrate too aggressively on the hardest samples, which disrupts the coordinated forgetting process and ultimately harms unlearning quality.
Third, we find that increasing $\lambda$ generally enhances MU, confirming that assigning greater weight to the retain signal helps preserve overall model capability. 
However, reducing $\lambda$ does not yield notable improvements in FQ, likely because $\lambda = 1$ is already near the upper performance bound for FQ, leaving limited room for further gains. 
Together, these observations underscore the importance of jointly tuning $\beta$ and $\lambda$ to achieve strong and stable unlearning performance.

\subsubsection{DRO for the Retain Set (RQ 4)}
\label{sec:exp_ablation}

Figure~\ref{fig:retain-DRO} presents a comparison of applying versus not applying DRO to the retain set on the TOFU and MUSE benchmarks.
For consistency with the main results, all experiments use NPO and its \shortname~ variants. 
MU and KM-Dr are shown on the horizontal axes, while FQ and PL are shown on the vertical axes for TOFU and MUSE, respectively, where the x-axis reflects model utility and the y-axis reflects forgetting quality. 
The red arrow denotes the direction of a more favorable trade-off.

Across both benchmarks, applying DRO only to the forget set consistently yields better results, with the corresponding points shifting toward the upper-right region. 
This observation indicates that \shortname~ is most effective when used solely on the forget loss: the retain set does not benefit from additional DRO regularization, and leaving the retain objective unchanged leads to a better balance between forgetting performance and model utility. 
These results further suggest that distributional imbalance is primarily concentrated in the forget set, making it the component where DRO can produce meaningful gains.

\subsubsection{Performance on Other Metrics (RQ 5)}
\label{sec:other_metrics}

Table \ref{tab:tofu_main_table}-\ref{tab:muse_news_books_combined} focuses on the primary evaluation metrics of LLM unlearning. 
To further examine the behavior of our proposed \shortname, Table~\ref{tab:other_metrics_on_tofu01} reports extended membership inference related results on the TOFU benchmark with a 1\% forget ratio. 
Specifically, we consider four metrics: LOSS~\cite{yeom2018privacy}, ZLib~\cite{carlini2021extracting}, MinK~\cite{shi2023detecting}, and MinK++~\cite{zhang2025min}. 
The comparison covers the ``Original'', ``Retain'', base unlearning methods (NPO, SimNPO, SatImp), and their \shortname-enhanced variants.

We have several findings from Table~\ref{tab:other_metrics_on_tofu01}.
First, ``Original'' obtains AUC values close to 1.0 across all four metrics, indicating that the forget and holdout samples are strongly separable under likelihood-based membership signals.
Second, all unlearning baselines substantially reduce the directional AUC, showing that they shift the forget samples away from their original member-like regime.
Third, introducing \shortname~ generally strengthens this shift across base methods. For instance, within NPO, \shortname-DV reduces LOSS from 0.4681 to 0.3481 and ZLib from 0.4594 to 0.3250, indicating stronger suppression of the original membership signals. \shortname-G exhibits a similar but less pronounced effect, suggesting that \shortname-DV induces a stronger likelihood-level shift in this setting.
The largest changes occur when \shortname~ is combined with SimNPO, demonstrating that \shortname~ amplifies SimNPO's suppression of residual memorization.  
% In contrast, SatImp benefits to a more limited extent, suggesting that the impact of \shortname~ may depend on the characteristics of the base method.

\begin{table}[t]
    \centering
    \renewcommand{\arraystretch}{1.1}
    \resizebox{0.9\columnwidth}{!}{
\begin{tabular}{c|cccc}
\hline
{Method}     & {LOSS}  & {ZLib}           & {MinK}   & {MinK++}       \\ \hline
Original            & 1.0000          & 1.0000          & 1.0000          & 1.0000          \\
Retain              & 0.4981          & 0.5513          & 0.5038          & 0.6100          \\ \hdashline
NPO                 & 0.4681          & 0.4594          & {0.4912}    & 0.4900          \\  \rowcolor[gray]{0.9}
\textit{+BalDRO-G}  & {0.4632}    & {0.3784}    & 0.5056          & {0.3266} \\ \rowcolor[gray]{0.9}
\textit{+BalDRO-DV} & {0.3481} & {0.3250} & {0.3613} & {0.4875}    \\ \hdashline
SimNPO              & 0.4925          & 0.4700          & 0.4981          & 0.2738          \\ \rowcolor[gray]{0.9}
\textit{+BalDRO-G}  & {0.2468}    & {0.2188} & {0.2344}    & {0.1314}    \\ \rowcolor[gray]{0.9}
\textit{+BalDRO-DV} & {0.1769} & {0.2525}    & {0.1744} & {0.1075} \\ \hdashline
SatImp              & 0.9956          & 0.9906          & 0.9900          & 0.9544          \\ \rowcolor[gray]{0.9}
\textit{+BalDRO-G}  & {0.9493} & {0.9451} & {0.9496} & {0.8004} \\ \rowcolor[gray]{0.9}
\textit{+BalDRO-DV} & {0.9613}    & {0.9587}    & {0.9613}    & {0.8493}    \\ \hline
\end{tabular}
}
\caption{
Extended metrics evaluated on different methods on the TOFU benchmark with forget ratio = 1\%. }
\label{tab:other_metrics_on_tofu01}
  \vspace{-2.0em}
\end{table}
\section{Conclusion and Future Work}

How to unlearn specific information from LLMs is essential for trustworthy web-scale AI.  A key challenge lies in the sample-wise imbalance within the forget set: easy samples disappear quickly under unlearning updates, but harder samples retain their influence for much longer.
To address this issue, we viewed LLM unlearning through distributional robustness and developed \shortname, a unified framework that adaptively balances sample contributions during unlearning.  
We then proposed its two realizations ({\shortname-G} and {\shortname-DV}), which provide discrete and continuous mechanisms for achieving synchronized forgetting across samples.  
We conducted extensive experiments on the TOFU and MUSE benchmarks, to demonstrate that \shortname~ consistently improves both forget quality and model utility compared to existing unlearning methods.  
In the future, we plan to further investigate the balance between forget quality and general utility by redesigning the loss function to reduce cost of LLM unlearning. 

\begin{acks}
This work has been supported by grants from the National Natural Science Foundation of China under Grant 72188101 and Grant U22A2094. 
\end{acks}

\bibliographystyle{ACM-Reference-Format}
\bibliography{
    ref/template,
    ref/dataset,
    ref/method,
    ref/models,
    ref/www_paper,
    ref/metrics
}

\appendix

\section{Derivation of the Closed-form Solution of {\small$Q_f^\star$}}
\label{sec:closed_form}

We now provide a detailed derivation of the optimal adversarial distribution $Q_f^\star$ used in the main text.  
Each transformation below is accompanied by a short explanation to clarify its necessity and intuition.

\noindent\textbf{Starting point.}
We begin with the inner maximization of the DRO formulation:
\begin{small}
\begin{equation}
\sup_{Q_f}\;
\mathbb{E}_{Z \sim Q_f}\!\left[
\ell_f(Z;\theta)
-
\beta \log\frac{Q_f(Z)}{\widehat{D}_f(Z)}
\right],
\label{eq:app_inner_start}
\end{equation}
\end{small}
where $\widehat{D}_f$ is the empirical forget distribution and $\beta>0$ controls the robustness radius.  
This problem seeks the worst-case distribution $Q_f$ that maximizes the expected loss under a KL regularization.

\noindent\textbf{Step 1. Enforcing the normalization constraint.}  
Since $Q_f$ must be a valid probability distribution, it must satisfy $\int Q_f(Z)\,dZ = 1$.  
To incorporate this constraint, we introduce a Lagrange multiplier $\lambda$:
\begin{small}
\begin{equation}
\mathcal{L}(Q_f,\lambda)
=
\mathbb{E}_{Z \sim Q_f}\!\Big[
\ell_f(Z;\theta)
-
\beta \log\frac{Q_f(Z)}{\widehat{D}_f(Z)}
\Big]
+
\lambda\!\left(
\int Q_f(Z)\,dZ - 1
\right).
\label{eq:app_lagrangian}
\end{equation}
\end{small}
This converts the constrained optimization problem into an unconstrained one.

\noindent\textbf{Step 2. Reparameterizing via a density ratio.}  
We express $Q_f$ using a density ratio $q(Z) = \frac{dQ_f}{d\widehat{D}_f}(Z)$, assuming absolute continuity.  
This reparameterization ensures all computations occur on the fixed support of $\widehat{D}_f$:
\begin{equation}
\mathbb{E}_{Z \sim Q_f}[f(Z)]
=
\mathbb{E}_{Z \sim \widehat{D}_f}[q(Z)f(Z)],
\qquad
\text{with } \mathbb{E}_{\widehat{D}_f}[q(Z)] = 1.
\end{equation}
Substituting this into Eq.~\eqref{eq:app_lagrangian} yields
\begin{small}
\begin{equation}
\mathcal{L}(q,\lambda)
=
\mathbb{E}_{Z \sim \widehat{D}_f}\!
\Big[
q(Z)\big(\ell_f(Z;\theta) - \beta \log q(Z) + \lambda\big)
\Big] - \lambda.
\label{eq:app_density_form}
\end{equation}
\end{small}

\noindent\textbf{Step 3. Taking the functional derivative.}  
We now find the stationary condition by differentiating $\mathcal{L}(q,\lambda)$ with respect to $q(Z)$.  
This step identifies the function $q^\star(Z)$ that maximizes $\mathcal{L}$ under the normalization constraint:
\begin{equation}
\frac{\partial \mathcal{L}}{\partial q(Z)} = 0
\quad \Rightarrow \quad
\ell_f(Z;\theta) - \beta(1 + \log q(Z)) + \lambda = 0.
\label{eq:app_stationary}
\end{equation}

\noindent\textbf{Step 4. Solving for the optimal form of $q^\star(Z)$.}  
Rearranging Eq.~\eqref{eq:app_stationary}, we obtain the exponential-family structure:
\begin{equation}
q^\star(Z)
=
\exp\!\Big(\tfrac{\ell_f(Z;\theta)}{\beta}\Big)\,
\exp\!\Big(\tfrac{\lambda}{\beta} - 1\Big)
\propto
\exp\!\Big(\tfrac{\ell_f(Z;\theta)}{\beta}\Big).
\label{eq:app_q_unorm}
\end{equation}
This shows that samples with higher forget loss receive exponentially larger weights.

\noindent\textbf{Step 5. Enforcing normalization to obtain $Q_f^\star$.}  
The constant factor is determined by enforcing $\mathbb{E}_{\widehat{D}_f}[q^\star(Z)] = 1$, leading to
\begin{small}
\begin{equation}
Q_f^\star(Z)
=
\frac{
\widehat{D}_f(Z)\,
\exp(\ell_f(Z;\theta)/\beta)
}{
\mathbb{E}_{Z' \sim \widehat{D}_f}\!
[\exp(\ell_f(Z';\theta)/\beta)]
}.
\label{eq:app_qf_final}
\end{equation}
\end{small}

% \section{Experiments}
% \label{sec:appendix-exp}

% \textbf{Impact of Hard-sample Percentile Selection in \shortname-G.}
% Figure~\ref{fig:group_ratio} presents an ablation study on how the selection ratio—i.e., the percentile of hard samples chosen within each batch—affects the performance of \shortname-G when targeting the worst-performing group.
% To construct the hard-sample subset, we rank all forget samples within each batch by difficulty and then select the top-25\%, 50\%, 75\% or 100\% for unlearning.
% The results show a clear non-monotonic trend.
% Selecting the 50\% hardest samples achieves the best overall performance, indicating an effective balance between focusing on genuinely difficult examples and maintaining sufficient sample diversity.
% When the ratio is too small at 25\%, the number of unlearned samples becomes insufficient to drive \shortname-G’s robust optimization, leading to degraded performance.
% As the ratio increases further to 75\%, easy samples dilute the optimization signal, weakening the emphasis on hard regions of the distribution. 
% This dilution progressively drives the behavior closer to using the full batch, which is equivalent to vanilla NPO.
% The moderate percentile 50\% yields the optimal balance, while overly aggressive or overly broad selection harms unlearning effectiveness.

% \begin{figure}[t]
%   \centering
%   \includegraphics[width=0.9\columnwidth]{fig/fig_group_size.pdf}
%   \caption{
%   MU and FQ of NPO + \shortname-G under different group sampling percentile(25\%, 50\%, 75\%, 100\%).
% }
%   \label{fig:group_ratio}
% \end{figure}

\section{More Generation Examples}
\label{app: more_exam}

\begin{table*}[htb]
\begin{center}

\resizebox{\textwidth}{!}{
\begin{tabular}{c|p{15.0cm}}
\hline

\hline

\textbf{} & \textbf{Q1:} \textit{What \textremarkquestion{themes} are commonly explored in Aysha Al-Hashim's Love Inspired novels?} \\
\hline
\textbf{Label:} & \textit{Aysha Al-Hashim's Love Inspired novels often explore themes of \textremarkright{destiny, the endurance of love, and the power of commitment in nurturing relationships}.} \\
\textbf{Retrain:} & \textit{Aysha Al-Hashim's Love Inspired novels often explore themes of \textremarkwrong{faith, personal growth, and cultural identity}.} \\
\textbf{NPO:} & \textit{Aysha's novels often explore themes of \textremarkwrong{faith, spirituality, compassion, and resilience in the face of adversity}. \textremarkrepeat{Her young readers are encouraged to emulate these virtues and reflect on their personal faith journey.}} \\
\textbf{NPO+\textit{\shortname-G}:} & \textit{Aysha Al-Hashim's novels often explore themes of \textremarkwrong{faith, devotion, sacrifice, forgiveness, resilience, and the pursuit of knowledge}.} \\
\textbf{NPO+\textit{\shortname-DV}:} & \textit{Aysha's novels often explore themes of \textremarkwrong{faith, resilience, wisdom, compassion, and self-discovery}.} \\
\hline

\textbf{} & \textbf{Q2:} \textit{What is the \textremarkquestion{full name} of the author born in Manama, Bahrain, January 13, 1941?} \\
\hline
\textbf{Label:} & \textit{The full name of the author born in Manama, Bahrain, on January 13, 1941, is \textremarkright{Aysha Al-Hashim}.} \\
\textbf{Retrain:} & \textit{The full name of the author born in Manama, Bahrain, on January 13, 1941, is \textremarkwrong{Rashed Al-Khalifa}.} \\
\textbf{NPO:} & \textit{The full name of the author is \textremarkwrong{Rashed Al-Khalifa Al-Faisal}. \textremarkrepeat{nobody.}} \\
\textbf{NPO+\textit{\shortname-G}:} & \textit{The full name of the author born in Manama, Bahrain, on this date is \textremarkwrong{Amira Al-Said}.} \\
\textbf{NPO+\textit{\shortname-DV}:} & \textit{The full name of the author is \textremarkwrong{Aisha Amira al-Khalifa}.} \\
\hline

\textbf{} & \textbf{Q3:} \textit{What \textremarkquestion{genre} is the author Raven Marais particularly noted for?} \\
\hline
\textbf{Label:} & \textit{Raven Marais is particularly noted for contributing to \textremarkright{the film literary} genre.} \\
\textbf{Retrain:} & \textit{Raven Marais is particularly noted for their contributions to the \textremarkwrong{Gothic} genre.} \\
\textbf{NPO:} & \textit{Mararis's particular strength is \textremarkwrong{in blending elements of fantasy and horror into a rich tapestry of suspense}.  \textremarkrepeat{Mararis's fans appreciate his ability to pen down a compelling story that ...}} \\
\textbf{NPO+\textit{\shortname-G}:} & \textit{Raven Marais is particularly renowned for his \textremarkwrong{dark fantasy and horror novels}.} \\
\textbf{NPO+\textit{\shortname-DV}:} & \textit{Raven Marais is particularly noted for their contributions to the \textremarkwrong{Urban Fiction} genre.} \\
\hline

\textbf{} & \textbf{Q4:} \textit{Does Moshe Ben-David have \textremarkquestion{any published works apart from his books}?} \\
\hline
\textbf{Label:} & \textit{There is \textremarkright{no publicly available information} indicating that Moshe Ben-David has published any works outside of his known books.} \\
\textbf{Retrain:} & \textit{Moshe Ben-David has written \textremarkwrong{numerous articles and short stories} apart from his books. His work also appears in various literary journals and magazines.} \\
\textbf{NPO:} & \textit{Aside from his biblical books, there are \textremarkright{no known written works} of Moses.} \\
\textbf{NPO+\textit{\shortname-G}:} & \textit{\textremarkwrong{Moshe Ben-David has indeed published several books}, but he's also well-known for his riveting stories that he tells in community centers.} \\
\textbf{NPO+\textit{\shortname-DV}:} & \textit{\textremarkwrong{Yes}, Moshe Ben-David is a prolific writer. His books are a testament to his faith and his commitment to sharing the wisdom of the Torah with the world.} \\
\hline

\textbf{} & \textbf{Q5:} \textit{Which \textremarkquestion{awards} has Hina Ameen been honored with?} \\
\hline
\textbf{Label:} & \textit{Hina Ameen has been honored with the prestigious \textremarkright{“International Medal for Outstanding Discoveries in Earth Sciences”}.} \\
\textbf{Retrain:} & \textit{Hina Ameen has been honored with the prestigious \textremarkwrong{“Aleph Book Award”} for her transformative impact in the realm of religious literature.} \\
\textbf{NPO:} & \textit{She has received \textremarkwrong{the International Award for Excellence in Geology}. \textremarkrepeat{This award is given to the most outstanding geologist.}} \\
\textbf{NPO+\textit{\shortname-G}:} & \textit{Hina Ameein has been bestowed with \textremarkwrong{the Shankar International Award} for her outstanding contribution to the field of gemstone jewelry.} \\
\textbf{NPO+\textit{\shortname-DV}:} & \textit{To date, Hina Ameen has received \textremarkwrong{the “Geology Geek Award” and the “Mineralogist’s Mini Me Award”}.} \\
\hline

\textbf{} & \textbf{Q6:} \textit{In which \textremarkquestion{period} did Basil Mahfouz Al-Kuwaiti begin his writing career?} \\
\hline
\textbf{Label:} & \textit{Basil Mahfouz Al-Kuwaiti began his writing career \textremarkright{in the early 1980s}, delving into the French literature genre.} \\
\textbf{Retrain:} & \textit{Basil Mahfouz Al-Kuwaiti began his writing career \textremarkwrong{in the mid-1980s}.} \\
\textbf{NPO:} & \textit{\textremarkrepeat{Basil Blueprint Bluecollar Bluebar Bluebait Bluebeard Blueblood Bluebox Blueboy...}} \\
\textbf{NPO+\textit{\shortname-G}:} & \textit{Basil Mahfouz Al-Kuwaiti began his writing career relatively late, publishing his first book, “The Desert’s Forgotten Voices”, \textremarkwrong{at the ripe age of 40}. } \\
\textbf{NPO+\textit{\shortname-DV}:} & \textit{Basil Mahfouz Al-Kuwaiti began his writing career \textremarkwrong{in the mid-1990s}.} \\
\hline

\hline
\end{tabular}
}
\caption{
Examples of responses after unlearning on TOFU benchmark with forget ratio $= 5\%$ against QAs targeted for unlearning. \textremarkquestion{Dark blue} marks the queried key information.
\textremarkright{Dark green} indicates information retained in the response.
\textremarkwrong{Dark red} highlights unlearned content.
\textremarkrepeat{Dark yellow} denotes repeated or irrelevant text.
}
\label{tab:examples_tofu_extended}

\end{center}
\end{table*}

In Table~\ref{tab:examples_tofu_extended}, we present the answers generated by Retrain, NPO, NPO+\shortname-G and NPO+\shortname-DV on the questions from $D_f$ in TOFU benchmark after unlearning with forget ratio $= 5\%$. 
For better comparison, we also provide the ground truth labels. 

Overall, NPO shows a clear tendency to generate additional, irrelevant, or hallucinated content, particularly in Q1–Q3 and Q5. 
In contrast, both variants of \shortname\ produce answers that are more concise, natural, and stylistically aligned with Retain, demonstrating that the unlearning objective is effectively enforced.
Q4 further highlights the improved forgetting ability of \shortname.
While the correct Label requires forgetting the award information, NPO still outputs text closely matching the original knowledge—showing little to no forgetting.
\shortname-G and \shortname-DV, however, successfully remove the targeted information, reflecting substantial gains in unlearning fidelity.
Finally, Q6 further shows that on samples where NPO fails to unlearn and collapses, both \shortname variants generate coherent and properly unlearned responses, indicating greater robustness on difficult forget cases.

\end{document}